\RequirePackage{fix-cm}
\documentclass[a4paper,11pt]{article}

\usepackage{authblk}
\usepackage{amsfonts,amsmath,amssymb,amsthm}
\usepackage{algorithm,algorithmic}
\usepackage{paralist}
\usepackage{multirow}
\usepackage{booktabs}
\usepackage{url}
\usepackage{hyperref}
\usepackage{graphicx}
\usepackage{microtype}
\usepackage{url}
\usepackage{hyperref}
\usepackage{subcaption}
\usepackage{tikz}
\usetikzlibrary{positioning}
\usetikzlibrary{decorations.pathreplacing}
\newcommand{\N}{\mathbb{N}}

\newcommand{\F}{\mathbb{F}}

\tikzset{
    mybrace/.style={decorate,decoration={brace,aspect=#1}}
}

\providecommand{\keywords}[1]{\textbf{\textit{Keywords }} #1}

\begin{document}

\title{Evolutionary Construction of Perfectly Balanced Boolean Functions}

\author[1]{Luca Mariot}
\author[1]{Stjepan Picek}
\author[2]{Domagoj Jakobovic}
\author[2]{Marko Djurasevic}
\author[3]{Alberto Leporati}

\affil[1]{{\small Digital Security Group, Radboud University, PO Box 9010, 6500 GL Nijmegen, The Netherlands} 
	
	{\small \texttt{\{luca.mariot,stjepan.picek\}@ru.nl}}}

\affil[2]{{\small Faculty of Electrical Engineering and Computing, University of Zagreb, Unska 3, Zagreb, Croatia}
	
	{\small \texttt{\{domagoj.jakobovic,marko.durasevic\}@fer.hr}}}

\affil[3]{{\small Dipartimento di Informatica, Sistemistica e Comunicazione, Università degli Studi di Milano-Bicocca, Viale Sarca 336/14, Milano, 20126, Italy}
    	{\small \texttt{alberto.leporati@unimib.it}}}

\maketitle

\begin{abstract}
Finding Boolean functions suitable for cryptographic primitives is a complex combinatorial optimization problem, since they must satisfy several properties to resist cryptanalytic attacks, and the space is very large, which grows super exponentially with the number of input variables. Recent research has focused on the study of Boolean functions that satisfy properties on restricted sets of inputs due to their importance in the development of the FLIP stream cipher. In this paper, we consider one such property, perfect balancedness, and investigate the use of Genetic Programming (GP) and Genetic Algorithms (GA) to construct Boolean functions that satisfy this property along with a good nonlinearity profile. We formulate the related optimization problem and define two encodings for the candidate solutions, namely the truth table and the weightwise balanced representations. Somewhat surprisingly, the results show that GA with the weightwise balanced representation outperforms GP with the classical truth table phenotype in finding highly nonlinear WPB functions. This finding is in stark contrast to previous findings on the evolution of globally balanced Boolean functions, where GP always performs best.
\end{abstract}


\section{Introduction}
\label{sec:intro}

Boolean functions are mathematical objects that are used in various domains like combinatorial design~\cite{Rothaus,954512}, coding theory~\cite{4167738,KERDOCK1972182}, cryptography~\cite{Adams1997}, and telecommunications~\cite{Paterson}. 
Since Boolean functions are widely used, it stands to reason that the design requirements also differ, necessitating diverse construction options.
Commonly, Boolean functions are constructed by following some algebraic construction, random search, or heuristics.
Comparing those approaches, we can conclude that heuristics are uniquely positioned as they allow the search for Boolean functions with any specific properties. Naturally, whether such an approach will yield an acceptable solution is difficult to predict as it depends on the problem difficulty (the combinations of properties that Boolean functions need to fulfill) and the size of Boolean functions that should be used.
One well-explored family of techniques to design Boolean functions are evolutionary algorithms (EAs).

Recently, Boolean functions started to be used in a new interesting scenario in cryptography, requiring them to be restricted over subsets of input vectors. Such Boolean functions can then be used in modern stream ciphers like FLIP~\cite{Meaux2016}. In this context, we are interested in Boolean functions of $n$ variables where all restrictions of Hamming weights between 1 and $n-1$ are balanced (i.e., have the same number of zeros and ones). Such functions are called Weightwise Perfectly Balanced (WPB) Boolean functions.

In general, finding Boolean functions that fulfill specific properties is not easy if they depend on a large number of inputs. Since the search space for $n$-variable Boolean functions is $2^{2^n}$, an exhaustive search is already impossible for $n=6$. Furthermore, while it is known that WPB Boolean functions need to have a required number of monomials in their algebraic normal form, there are only a few known algebraic constructions for WPB functions. Thus, heuristic algorithms provide an interesting perspective to construct this kind of functions.

Finding WPB functions is a relevant problem from the algebraic perspective as such functions are relatively ``new'' and cryptographic perspective as they can be used in specific types of stream ciphers. What is more, due to limited results up to now, it is also not known what the best possible cryptographic properties are (e.g., nonlinearity) that such functions can reach. Indeed, the currently known bounds are rather loose, and improvements could be possible. Finding them with evolutionary algorithms would prove that such techniques have their place even in the domains traditionally reserved for more deterministic approaches.

To the best of our knowledge, no existing works use heuristics to design WPB functions. Still, if we consider the body of works (see Section~\ref{sec:related}) that use evolutionary algorithms to evolve Boolean functions with various properties, it provides us with optimism that interesting results are possible.

This work concentrates on evolving WPB Boolean functions with eight inputs. To this end, we use a genetic algorithm (GA) and a genetic programming (GP) technique. More precisely, we start by analyzing the main difficulties in such a process and discuss how evaluating the nonlinearity property becomes a bottleneck as it is not possible to use efficient algorithms. Then, we discuss how calculation can still be improved by restricting our attention to only part of the calculation. Finally, we experiment with different variants of evolutionary algorithms, where for some we impose requirements that the solutions must be balanced to reduce the search space size and make the optimization process more efficient.

Our results show that both GA and GP can reach a good performance on this problem, and moreover GA based on the weightwise balanced representation can even surpass GP with the classic tree-to-table encoding. This represents a remarkable result as it gives us Boolean functions with good restricted nonlinearities, but also becomes one of the rare scenarios where GA outperforms GP when evolving Boolean functions.

\section{Background}
\label{sec:background}
Here we recall the basic definitions and results related to Boolean functions and their cryptographic properties used in the remainder of the paper. We refer the reader to Carlet's book~\cite{carlet2021} for a more thorough treatment of the subject. 

Let $\F_2 = \{0,1\}$ be the finite field with two elements, with the sum (i.e., XOR) and multiplication (i.e., AND) operations respectively denoted by $\oplus$ and concatenation. For any $n \in \N$, we denote by $\F_2^n$ the $n$-dimensional vector space over $\F_2$, with the vector sum defined coordinate-wise. Given $x, y \in \F_2^n$, their \emph{dot product} is given by $x \cdot y = \bigoplus_{i=1}^n x_iy_i$. The Hamming distance $d_H(x,y)$ of $x$ and $y$ is the number of coordinates in which they differ. The Hamming weight $w_H(x)$ of a vector~$x$ is the Hamming distance between $x$ and the null vector $\underline{0}$. Equivalently, the Hamming weight of vector~$x$ can be defined as the cardinality of the \emph{support} of $x$, that is, $w_H(x) = |supp(x)| = |\{i : x_i \neq 0\}|$.

An $n$-variable \emph{Boolean function} is a mapping $f: \F_2^n \to \F_2$. The most common way to uniquely represent $f$ is through its \emph{truth table}, which is the $2^n$-bit vector specifying the values of $f$ for all possible input vectors in $\F_2^n$, assuming they are lexicographically ordered. The Hamming weight of $f$ is then simply defined as the weight of its truth table, and it is denoted as $w_H(f)$; further, $f$ is \emph{balanced} if $w_H(f) = 2^{n-1}$, or equivalently if its truth table is composed of an equal number of zeros and ones. Balancedness is a fundamental cryptographic criterion for Boolean functions used in stream and block ciphers designs: indeed, unbalanced functions have a statistical bias that can be exploited in attacks~\cite{carlet2021}.

The \emph{Algebraic Normal Form} (ANF) is another common method used in cryptography to uniquely represent a Boolean function. Given $f: \F_2^n \to \F_2$, and observing that each element in $\F_2$ is \emph{idempotent} (that is, $x^2 = x$ for all $x \in \F_2$), the ANF of $f$ is defined as the following multivariate polynomial in the quotient ring $\mathbb{F}_2[x_1,\cdots,x_n]/(x_1^2 \oplus x_1, \cdots, x_n^2 \oplus x_n)$:
\begin{equation}
    P_f(x) = \bigoplus_{I \in 2^{[n]}} a_I \left( \prod_{i \in I} x_i \right) \enspace ,
\end{equation}
where $2^{[n]}$ denotes the power set of $[n] = \{1,\cdots,n\}$. The coefficients $a_I \in \F_2^n$ that determine the ANF polynomial can be recovered from the truth table of $f$ via \emph{M\"{obius} inversion}:
\begin{equation}
  \label{eq:mobius}
  a_I = \bigoplus_{x \in \F_2^n: supp(x) \subseteq I} f(x) \enspace .
\end{equation}
Then, the \emph{algebraic degree} of $f$ is defined as the largest monomial in the ANF of $f$, or equivalently as the cardinality of the largest $I \in 2^{[n]}$ such that $a_I \neq  0$.

Boolean functions of degree at most $1$ are also called \emph{affine functions}. Remark that the ANF of an affine function is basically an XOR of a subset of the input variables and a constant, or equivalently $a\cdot x \oplus b$ with $a \in \F_2^n$ and $b \in \F_2$. When $b=0$, the resulting function $a\cdot x$ is also called \emph{linear}. The \emph{nonlinearity} of a Boolean function $f: \F_2^n \to \F_2$ is the minimum Hamming distance of $f$ from the set of all $n$-variable affine functions. This criterion is very important in symmetric ciphers. Indeed, Boolean functions with a high nonlinearity (or equivalently, that are hard to approximate by affine functions) have better resistance towards fast-correlation attacks in stream ciphers and linear cryptanalysis in block ciphers.

The \emph{Walsh-Hadamard Transform} can be used to determine the nonlinearity of a Boolean function. Given $f: \F_2^n \to \F_2$ and $a \in \F_2^n$, the corresponding Walsh-Hadamard coefficient is defined as:
\begin{equation}
    \label{eq:walsh}
    W_f(a) = \sum_{x \in \F_2^n} (-1)^{f(x) \oplus a\cdot x} \enspace .
\end{equation}
In other words, $W_f(a)$ measures the correlation between $f$ and the linear function $a\cdot x$. The nonlinearity of $f$ is then:
\begin{equation}
\label{eq:nl}
nl(f) = 2^{n-1} - \frac{1}{2}\cdot \max_{a \in \F_2^n} |W_f(a)| \enspace .
\end{equation}
The nonlinearity of any Boolean function is bounded above by the inequality $nl(f) \le 2^{n-1} - 2^{\frac{n}{2}-1}$, which corresponds to the \emph{covering radius bound} for the first-order Reed-Muller code $(1,n)$. Such bound is tight only for $n$ even, and the functions which satisfy it are called \emph{bent}. Although they reach the highest possible nonlinearity, bent functions are also unbalanced and therefore unsuitable for cryptographic purposes. Determining the maximum nonlinearity is an open problem for any odd number of variables $n>7$.

Recently, the research on cryptographic properties of Boolean functions restricted over subsets of input vectors gained prominence, especially within the context of the FLIP stream cipher~\cite{carlet17}. Subsets of particular interests are those collecting all input vectors in $\F_2^n$ of a fixed Hamming weight $k$, defined as $E_{n,k} = \{x \in \F_2^n: \ w_H(x) = k\}$ for $k \in [n]$. The cardinality of $E_{n,k}$ is $\binom{n}{k}$, since it corresponds to the number of ways one can set $k$ ones in an $n$-bit string. The restriction of $f: \F_2^n \to \F_2$ to $E_{n,k}$ is denoted by $f_{(k)}$. The function $f$ is called \emph{Weightwise Perfectly Balanced} (WPB) if all restrictions $f_{(k)}$ of weight between 1 and $n-1$ are balanced:
\begin{equation}
w_H(f_{(k)}) = \frac{|E|}{2} = \frac{1}{2} \cdot \binom{n}{k} \enspace ,
\end{equation}
for all $k \in \{1,\cdots,n-1\}$. Obviously, $k=0$ and $k=n$ are excluded since $\binom{n}{0} = \binom{n}{n} = 1$. Furthermore, to obtain a function that is also globally balanced, one needs to impose the constraint that $f(\underline{0}) \neq f(\underline{1})$. In what follows, we will assume that $f(\underline{0}) = 0$ and $f(\underline{1}) = 1$.

A function $f: \F_2^n \to \F_2$ is WPB if and only if $n$ is a power of 2. This is a consequence of \emph{Lucas's theorem}, which states that  $\binom{n}{k} \equiv 1 \textrm{ mod  } 2$ if and only if $bin(k) \preceq bin(n)$, i.e., if and only if $bin(k)_i \le bin(n)_i$ for all positions $i$ in the binary expansions $bin(n)$ and $bin(k)$ of $n$ and $k$. The WPB condition can be relaxed by imposing that each restriction $f_{(k)}$ has Hamming weight $\frac{1}{2} \left(\binom{n}{k} \pm 1 \right)$ and $\frac{1}{2} \binom{n}{k}$ respectively when $\binom{n}{k}$ is odd and even. Functions satisfying this condition are also called \emph{Weightwise Almost Perfectly Balanced} (WAPB), and they exist also when $n$ is not a power of $2$. However, in this paper we consider only the perfectly balanced case. Hence, in what follows we assume that $n=2^m$ for $m \in \N$.

The nonlinearity property is straightforwardly adapted to the case of restricted inputs with a fixed Hamming weight. The Walsh-Hadamard transform of $f: \F_2^n \to \F_2$ over $E_{n,k}$ is defined for all $a \in \F_2^n$ as:
\begin{equation}
    \label{eq:walsh-restr}
    W_{f_{(k)}}(a) = \sum_{x \in E_{n,k}} (-1)^{f(x) \oplus a\cdot x} \enspace .
\end{equation}
Remark that the only difference between Equations~\eqref{eq:walsh} and~\eqref{eq:walsh-restr} is that, in the latter, the sum ranges over $E_{n,k}$ instead of $\F_2^{n}$. Similarly, the restricted nonlinearity of $f$ is defined in terms of the coefficients $W_{f_{(k)}}(a)$:
\begin{equation}
\label{eq:nl-k}
nl_k(f) = 2^{n-1} - \frac{1}{2}\cdot \max_{a \in \F_2^n} |W_{f_{(k)}}(a)| \enspace .
\end{equation}
Remark that, in particular, the maximum absolute value in Equation~\eqref{eq:nl-k} is considered among \emph{all} coefficients $a \in \F_2^n$, not only those of weight $k$. In other words, $nl_k(f)$ measures the distance between the restriction of $f$ to inputs of weight $k$ and all affine functions of $n$ variables. The analogous version of the covering radius bound for $nl_k$ is:
\begin{equation}
\label{eq:crb-k}
nl_k(f) \le \frac{1}{2}\cdot \binom{n}{k} - \frac{1}{2} \cdot \sqrt{\binom{n}{k}} \enspace .
\end{equation}
Carlet et al.~\cite{carlet17} showed that even when $\binom{n}{k}$ is a square, the above bound is not tight, and observed that it is an open question to determine when the floor of the right-hand side of Eq.~\eqref{eq:crb-k} can be satisfied with equality. Mesnager et al. proved a tighter bound in~\cite{mesnager19} and claimed that, in general, it could be much lower than the covering radius bound analog.

We conclude this section with a consideration on the algebraic normal forms of WPB functions. Carlet et al.~\cite{carlet17} proved that if $f: \F_2^n \to \F_2$ is WPB and $n>4$, then its ANF must be made of at least $\frac{3}{4}n+1$ monomials. The authors also remark that it is not known whether this is the smallest number possible since, in that work, they managed to find a construction of a WPB function with $n-1$ monomials.

As an example, Table~\ref{tab:ex-wpb} reports a WPB function of $n=4$ variables. The first two columns correspond to the global truth table of the function, with input vectors listed in lexicographic order. The remaining three columns respectively give the Hamming weights $0 \le k \le 4$, the input vectors in $E_{n,k}$ (again in lexicographic order), and the output value of the corresponding restriction $f_{(k)}$.

\begin{table}[t]
\centering
\caption{Example of a WPB function of $n=4$ variables.}
\begin{tabular}{ccccc}
\toprule
$\F_2^4$ & $f(x)$ & $k$ & $E_{4,k}$ & $f_{(k)}$ \\
\midrule
$(0,0,0,0)$ & $0$ & $0$ & $(0,0,0,0)$ & $0$ \\
\noalign{\smallskip}
\cline{3-5}
\noalign{\smallskip}
$(0,0,0,1)$ & $1$ & \multirow{4}{*}{$1$} & $(0,0,0,1)$ & $1$ \\
$(0,0,1,0)$ & $0$ &                      & $(0,0,1,0)$ & $0$ \\
$(0,0,1,1)$ & $1$ &                      & $(0,1,0,0)$ & $0$ \\
$(0,1,0,0)$ & $0$ &                      & $(1,0,0,0)$ & $1$ \\
\noalign{\smallskip}
\cline{3-5}
\noalign{\smallskip}
$(0,1,0,1)$ & $0$ & \multirow{6}{*}{$2$} & $(0,0,1,1)$ & $1$ \\
$(0,1,1,0)$ & $1$ &                      & $(0,1,0,1)$ & $0$ \\
$(0,1,1,1)$ & $0$ &                      & $(0,1,1,0)$ & $1$ \\
$(1,0,0,0)$ & $1$ &                      & $(1,0,0,1)$ & $0$ \\
$(1,0,0,1)$ & $0$ &                      & $(1,0,1,0)$ & $1$ \\
$(1,0,1,0)$ & $1$ &                      & $(1,1,0,0)$ & $0$ \\
\noalign{\smallskip}
\cline{3-5}
\noalign{\smallskip}
$(1,0,1,1)$ & $0$ & \multirow{4}{*}{$3$} & $(0,1,1,1)$ & $1$ \\
$(1,1,0,0)$ & $1$ &                      & $(1,0,1,1)$ & $0$ \\
$(1,1,0,1)$ & $1$ &                      & $(1,1,0,1)$ & $0$ \\
$(1,1,1,0)$ & $1$ &                      & $(1,1,1,0)$ & $1$ \\
\noalign{\smallskip}
\cline{3-5}
\noalign{\smallskip}
$(1,1,1,1)$ & $1$ & $4$ & $(1,1,1,1)$ & $1$ \\
\bottomrule
\end{tabular}
\label{tab:ex-wpb}
\end{table}

\section{Related Works}
\label{sec:related}

Evolutionary algorithms have been used to evolve Boolean functions with specific cryptographic properties for more than two decades already.
While there is an abundance of works in the literature, we can consider an algorithmic perspective and an objective perspective. From the objective side, the two dominant goals are to evolve either bent Boolean functions or balanced Boolean functions -- with high nonlinearity and, possibly, some additional cryptographic properties. From the algorithmic perspective, most research works either use genetic algorithms (GA) or genetic programming (GP).

The first paper investigating the evolutionary algorithms approach for the evolution of Boolean functions with cryptographic properties was published by Millan et al. ~\cite{Millan97}. There, the authors used GA to evolve Boolean functions with high nonlinearity.
Expanding on the previous results, Millan et al. used GA, hill climbing, and a resetting step to evolve highly nonlinear balanced Boolean functions with up to 12 inputs~\cite{millan}. Mariot and Leporati~\cite{mariot15} investigated the spectral inversion approach, originally pioneered by Clark et al.~\cite{clark04}, by proposing a GA to evolve Walsh spectra of pseudo-Boolean functions satisfying good cryptographic properties.
Picek and Jakobovic considered a different approach where instead of evolving Boolean functions, they evolved secondary constructions of Boolean functions~\cite{PicekJ16}. 
Jakobovic et al. investigated the difficulty of evolving Boolean functions with specific properties where they used fitness landscape analysis based on Local Optima Networks~\cite{DBLP:journals/asc/JakobovicPMW21}. 

The works discussed up to now considered ``classical'' Boolean functions. There are, however, also some works that consider different types of Boolean functions. Picek et al. investigated how to evolve quaternary bent Boolean functions~\cite{8477677}. There, instead of using the common binary case \{0,1\}, both the domain and the output of the truth table representation range over four possible values, namely \{0, 1, 2, 3\}.
Finally, Mariot et al. experimented with hyper-bent Boolean functions, which represent a significantly more difficult task to evolve than bent Boolean functions~\cite{DBLP:conf/eurogp/MariotJLP19}.

\section{Methodology}
\label{sec:description}

This section delves into the details of the evolutionary algorithms that we used to search for WPB functions, namely GA and GP. We start with some considerations on the search space underlying the optimization problem and then introduce the encodings for the candidate solutions, as well as the variation operators used to generate them. We then define the fitness functions that drive the search of GA and GP.

\subsection{Search Space Analysis}
\label{subsec:encoding}
The most straightforward way to search for WPB functions is to explore the whole space of $n$-variable Boolean functions $\mathcal{F}_{n} = \{f: \F_2^n \to \F_2\}$. A basic combinatorial argument shows that the size of this set is super-exponential in $n$. Indeed, each function is uniquely identified by its truth table, which is a vector of $2^n$ bits; therefore, $\mathcal{F}_n$ is composed of $2^{2^n}$ functions. Exhaustive enumeration of all solutions in $\mathcal{F}_n$ becomes unfeasible already for $n>5$. Since WPB functions exist only when $n$ is a power of $2$, the only two instances where WPB functions can be exhaustively searched are $n=2$ and $n=4$, which are too small for any interesting application or for obtaining theoretical insights about their structure. As an example, for $n=4$ there are $720$ WPB functions, 288 of which are linear (i.e., all their weightwise restrictions are linear), while the remaining ones all have $nl_1 = nl_3 = 0$ and $nl_2 = 1$. The next interesting instance is thus $n=8$, where the search space is composed of $2^{2^8} \approx 1.16 \cdot 10^{77}$ Boolean functions, clearly beyond reach for any attempt of exhaustive search. This basic remark is the first motivation to employ EAs for studying the structure of WPB functions.

A first refinement is to consider only the space of balanced Boolean functions, which we denote by $\mathcal{B}_n$ for all $n \in \N$. Since we impose $f(\underline{0}) = 0$ and $f(\underline{1}) = 0$ on the WPB functions to ensure that their truth tables are globally balanced, it makes sense to restrict the search space to $\mathcal{B}_n$, whose cardinality is $\binom{2^n}{2^{n-1}}$ for all $n \in \N$. However, the gain from this reduction is not very significant, since for $n=8$ one has that $\binom{256}{128} \approx 5.77 \cdot 10^{76}$. Moreover, this set still considers candidate solutions that are not WPB functions.

Taking the approach above further, we can derive a counting formula for the set of all WPB functions of $n \in \N$ variables, denoted by $\mathcal{W}_n$ in what follows. As explained in Section~\ref{sec:background}, for each weight $1 \le k \le n-1$ the space $E_{n,k}$ is composed of $\binom{n}{k}$ input vectors. Since the restricted truth table over this set must be balanced, it follows that we can choose how to set the $\frac{1}{2}\binom{n}{k}$ ones in it in the following number of ways:
\begin{equation}
\label{eq:binom-bal}
\mathcal{B}_{(n,k)} = \binom{\binom{n}{k}}{\frac{1}{2}\cdot\binom{n}{k}} \enspace .
\end{equation}
Observing that the truth table of each restriction $f_{(k)}$ is independent from the others, we finally obtain the number of WPB functions of $n$ variables:
\begin{equation}
\label{eq:wpb-num}
\mathcal{W}_n = \prod_{k=1}^{n-1} \binom{\binom{n}{k}}{\frac{1}{2}\cdot\binom{n}{k}} \enspace .
\end{equation}
For $n=8$, one thus obtains a search space of $\mathcal{W}_n \approx 5.18 \cdot 10^{70}$ elements in total, which is slightly better by a few orders of magnitudes than the cardinalities of $\mathcal{F}_n$ and $\mathcal{B}_n$. Further, the search process explores only WPB functions, which allows one to focus the optimization effort on the restricted nonlinearities.

Table~\ref{tab:sizes} compares the sizes of the search spaces of all Boolean functions ($\#\mathcal{F}_n$), balanced functions ($\#\mathcal{B}_n$) and WPB functions ($\#\mathcal{W}_n$) up to $n=16$.
\begin{table}[t]
    \caption{(Approximate) search space sizes for various $n$.}
    \label{tab:sizes}
    \centering
    \begin{tabular}{clll}
    \toprule
    $n$ & $\#\mathcal{F}_n$ & $\#\mathcal{B}_n$ & $\#\mathcal{W}_n$ \\
    \midrule
    $2$ & $16$ & $6$ & $2$ \\
    $4$ & $65536$ & $12870$ & $720$ \\
    $8$ & $1.16 \cdot 10^{77}$ & $5.77 \cdot 10^{76}$ & $5.28 \cdot 10^{70}$ \\
    $16$ & $2.01 \cdot 10^{19729}$ & $6.24 \cdot 10^{19727}$ & $1.84 \cdot 10^{19704}$ \\
    \bottomrule
    \end{tabular}
\end{table}
Remark that the restricted Walsh-Hadamard transforms can be computed only in a \emph{naive} fashion, i.e., by iterating through all terms in the sum of Equation~\eqref{eq:walsh-restr}, which gives a quadratic complexity of $2^{2n}$ operations. Contrarily, with the general Walsh-Hadamard transform defined in Equation~\eqref{eq:walsh}, there exists a \emph{Fast Walsh Transform} (FWT) algorithm with a logarithmic complexity of $n2^n$ operations~\cite{carlet2021}. This explains why, in previous works on the evolution of Boolean functions, researchers could scale up to $n=16$ variables and beyond in their experiments. However, since the FWT algorithms depend on a divide-and-conquer strategy, it requires that the number of coefficients is a power of $2$, something which does not hold in general for the weightwise subsets $E_{n,k}$ over which the restricted Walsh transform is defined. For this reason, in our experiments, we considered only the problem instance with $n=8$ variables.

\subsection{Solutions Encoding and Variation Operators}
\label{subsec:variation}
In principle, each representation of the search space of interest can be used to formulate a proper encoding for the candidate solutions searched by EAs. In this paper, we considered two encodings stemming from the discussion in the previous section, namely the basic \emph{truth table representation} and the \emph{weightwise balanced representation}.

In the truth table representation, each candidate solution is encoded by a string of $2^n$ bits, corresponding to the truth table vector of an $n$-variable Boolean function. We adopted this representation to evolve WPB functions with a standard GA, using classic one-point crossover and flip mutation. Clearly, these operators do not preserve the Hamming weight of a Boolean function, let alone its restricted balancedness over the weightwise subsets $E_{n,k}$. Therefore, our GA based on this truth table representation searches in the whole set $\mathcal{F}_n$ of all $n$-variable Boolean functions.

We also adopted the truth table representation for our GP experiments, although a further encoding step is required in this case. Indeed, GP manipulates \emph{syntactic trees} instead of bitstrings. As usual in related works on GP and cryptographic Boolean functions, we represent a candidate solution by a tree whose leaf nodes correspond to the input variables $x_1,\cdots, x_n \in \F_2$. The internal nodes are Boolean operators that combine the inputs received from their children and propagate their output to the respective parent nodes. The Boolean functions used by the GP are OR, XOR, AND, AND2, XNOR, and function NOT that takes a single argument. The function AND2 behaves the same as the function AND but with the second input inverted, whereas the IF function takes three arguments and returns the second one if the first one evaluates to true and the third one otherwise. Thus, the output of the root node is the output value of the Boolean function. The corresponding truth table of the function $f: \F_2^n \to \F_2$ is determined by evaluating the tree over all possible $2^n$ assignments of the input variables at the leaf nodes. We then employed standard GP variation operators such as subtree crossover and subtree mutation, which are not generally weight-preserving. Hence, similarly to the standard GA with the truth table representation, our GP searches the whole set of $n$-variable Boolean functions $\mathcal{F}_n$.

Remark that with the truth table representation we always force $f(\underline{0}) = 0$ and $f(\underline{1}) = 1$ on the candidate solutions before evaluating their fitness, to ensure that the WPB functions evolved through GA and GP are also globally balanced.

Unlike the truth table, the weightwise balanced representation allows to \emph{search only inside the space of WPB functions} $\mathcal{W}_n$. This is accomplished as follows. Given the target number of variables $n \in \N$, we formally define the genotype of a candidate solution $C$ as:
\begin{equation}
    \label{eq:chrom-bal}
    C = \left \{c_{n,k} \in \F_2^{\binom{n}{k}}: \ 1 \le k \le n-1, \ w_H(c_{n,k}) = \frac{1}{2} \binom{n}{k} \right \} \enspace .
\end{equation}
In other words, $C$ is a set of $n-1$ bitstrings, where the length of each bitstring $c_{n,k}$ is the cardinality of the restricted subset $E_{n,k}$. Further, the Hamming weight of each $c_{n,k}$ is half of its length, which means that the restriction $f_{(k)}$ whose truth table is defined by $c_{n,k}$ is balanced. The phenotype corresponding to $C$ is the  function $f_C: \F_2^n \to \F_2$ defined as $f_C(\underline{0}) = 0$, $f_C(\underline{1}) = 1$, and $f(x) = c_{n,k}[i]$ for all $x \in \F_2^n$ such that $w_H(x) = k$ for $1 \le k \le n-1$. Here, $c_{n,k}[i]$ denotes the $i$-th bit of $c_{n,k}$, where $i$ corresponds to the position of $x$ in $E_{n,k}$ in lexicographic order. The last column of Table~\eqref{tab:ex-wpb} can be taken as an example of a weightwise balanced representation for $n=4$ variables; in this case, the chromosome is the following:
\begin{equation}
    \label{eq:chrom-bal-ex}
    C = \{1001, 101010, 1001\} \enspace .
\end{equation}

We employed the weightwise balanced representation only with GA, since in this case it is possible to define variations operators that preserve the Hamming weights of the candidate solutions. Manzoni et al.~\cite{manzoni20} performed a thorough statistical analysis of three balanced crossover operators over different combinatorial optimization problems, some of which related to the cryptographic properties of Boolean functions. For our experiments, we adopted the \emph{counter-based} (CB) and \emph{map-of-ones} (MO) balanced crossovers, which were found by the authors of~\cite{manzoni20} to have better performances over the \emph{zero-length} crossover. In our weightwise balanced representation, the CB and MO crossovers are applied \emph{independently} on each substring. Given two chromosomes $C = \{c_{n,1},\cdots c_{n,n-1} \}$ and $D = \{d_{n,1},\cdots d_{n,n-1} \}$, an offspring chromosome $O$ is obtained through CB (respectively, MO) crossover by defining $o_{n,k} = CB(c_{n,k},d_{n,k})$ (respectively, $o_{n,k} = MO(c_{n,k},d_{n,k})$) for all $1 \le k \le n-1$.

Concerning mutation, we employed a simple \emph{swap-based} operator that, for each position $i$ in $c_{n,k}$, exchanges with probability $p_{mut}$ the bit $c_{n,k}[i]$ with $c_{n,k}[j]$, where $j$ is chosen at random. The swap position $j$ must be selected inside $c_{n,k}$, to preserve the WPB property of the candidate solution.

\subsection{Fitness Functions}
\label{subsec:fitness}

To optimize the WPB property of Boolean functions, a possible way is to incorporate a penalty factor in the fitness that punishes unbalancedness over the subsets of fixed Hamming weight. The Hamming weight of the restriction $f_{(k)}$ is related to the Walsh-Hadamard coefficient of the null vector $W_{f_{(k)}}(\underline{0})$ via the following equation~\cite{carlet2021}:
\begin{equation}
\label{eq:hw-restr}
w_H(f_{(k)}) = \frac{\#E_{n,k} - W_{f_{(k)}}(\underline{0})}{2} \enspace ,
\end{equation}
where $\#$ denotes the cardinality of a set. This formula has the advantage that it is not necessary to evaluate the Hamming weight of each restriction separately from the Walsh-Hadamard spectrum, thus saving some computations. The unbalancedness of $f$ with respect to $E_{n,k}$ is then defined for all $k \in [n-1]$ as the deviation of its truth table from being balanced, that is:
\begin{equation}
\label{eq:unb-single}
unb_k(f) = \left | \frac{\#E_{n,k}}{2} - w_H(f_{(k)}) \right | \enspace .
\end{equation}
In other words, $unb_k(f)$ is the number of bits that need to be changed in the truth table of $f_{(k)}$ to make it balanced.

Clearly, the penalty factor for the WPB property must take into account the unbalancedness of $f$ with respect to all subsets of inputs. For this reason, we define it as the sum of all unbalancedness factors with $k \in [n-1]$:
\begin{equation}
\label{eq:penalty}
pen(f) = \sum_{k=1}^{n-1} unb_k(f) \enspace .
\end{equation}
Eq.~\eqref{eq:penalty} is always non-negative since it is a sum of absolute values. Therefore, one can subtract it in the fitness function to effectively minimize it.

The second property to be optimized are the nonlinearities over the restricted subsets $E_{n,k}$. For this part, we tested two different strategies:
\begin{compactenum}
\item[(a)] Maximize the \emph{sum} of the nonlinearities.
\item[(b)] Maximize the \emph{minimum} nonlinearity.
\end{compactenum}
Remark that Liu and Mesnager~\cite{liu19} showed that $nl_1(f) = 0$ for any WPB function $f$. The authors also proved that the restricted nonlinearities of WPB functions are symmetric with respect to the Hamming weight, that is, $nl_k(f) = nl_{n-k}(f)$ for $1 \le k \le \frac{n}{2}$. Hence, it is possible to optimize the computation of the nonlinearities for the two strategies above by taking into account only the weights $2 \le k \le n/2$. This is a considerable gain in evaluating the fitness functions since the restricted Walsh-Hadamard transforms can only be computed with the naive algorithm.

Hence, the two fitness functions maximized by our EAs are formally defined as follows:
\begin{align}
\label{eq:fit-1}
fit_1(f) &= \delta_{pen} \cdot \left( \sum_{k=2}^{n/2} nl_k(f) \right) - pen(f) \enspace , \\
\label{eq:fit-2}
fit_2(f) &= \delta_{pen} \cdot \left( \min_{2 \le k \le n/2} \{nl_k(f)\} \right) - pen(f) \enspace ,
\end{align}
where $\delta_{pen}$ is equal to $1$ when $pen(f) = 0$ (i.e., the function is WPB) and $0$ otherwise. This forces the EA first to optimize the unbalancedness penalty factor and then focus on the nonlinearities while retaining the WPB property, which allows considering only the weights between $2$ and $n/2$.

Obviously, the penalty factor can be safely omitted from the fitness functions when using GA with balanced variation operators since there the candidate solutions are always WPB.

\section{Experiments}
\label{sec:experiments}

This section presents the experimental evaluation of our approach to constructing WPB functions with EAs. We start by describing the experimental setting adopted for both GA and GP and then report the results obtained from our experiments.

\subsection{Experimental Settings}
\label{subsec:settings}
As remarked in Section~\ref{subsec:encoding}, we considered only Boolean functions of $n=8$ as a problem instance. Our GA and GP employed a steady-state operator with a 3-tournament elimination concerning the selection process. This means that, in each iteration, three individuals are chosen at random from the population for the tournament, and the worst one in terms of fitness value is eliminated. The other two remaining individuals in the tournament are used by the crossover operator to generate a new child individual, which then undergoes mutation with probability $p_{mut}$. Finally, the mutated child takes the place of the eliminated individual in the population.

We performed a preliminary tuning phase to select the best population sizes. In particular, for GA, we chose a population of 200 individuals, while for GP, the population was set to 1000 individuals. The mutation rate $p_{mut}$ for both GA and GP ranged in the set $\{0.1, 0.3, 0.5, 0.7, 0.9\}$. Further, the maximal depth for trees evolved by GP was set to 5. The genetic operators used by GP are simple tree crossover, uniform crossover, size fair, one-point, and context preserving crossover (selected at random at each crossover event) and subtree mutation~\cite{poli08:fieldguide}. On the other hand, for GA, we considered one-point crossover and flip mutation for the truth table representation, while counter-based, map-of-ones crossover, and swap-based mutation were used for the weightwise balanced representation~\cite{manzoni20}.

Further common experimental parameters include the number of fitness evaluations, which was set to 500000 since no improvements in both fitness functions were observed after that. Finally, each experiment was repeated for $30$ runs.

\subsection{Results}
\label{subsec:results}

In what follows, we denote by GA-OP the GA with truth table representation, and by GA-CB and GA-MO the GA with weightwise balanced representation respectively equipped with the counter-based and map-of-ones crossover operators. For each of the four considered EA variants (GA-OP, GA-CB, GA-MO, and GP), we recorded the fitness value of the best individual in the population at the end of each experimental run. The first remarkable finding is that \emph{there is no difference among the four EAs concerning the second fitness function $fit_2$}, i.e., the one that maximizes the minimum restricted nonlinearity and minimizes the WPB penalty factor when the truth table representation is used. As a matter of fact, all considered EAs achieved the same best fitness value of 10 across all runs. Therefore, in what follows, we focus on the results obtained with the first fitness function $fit_1$, which instead computes the sum of all restricted nonlinearities $nl_k$ for $k \in \{2,3,4\}$.

\begin{figure}[t]
\includegraphics[width=\columnwidth]{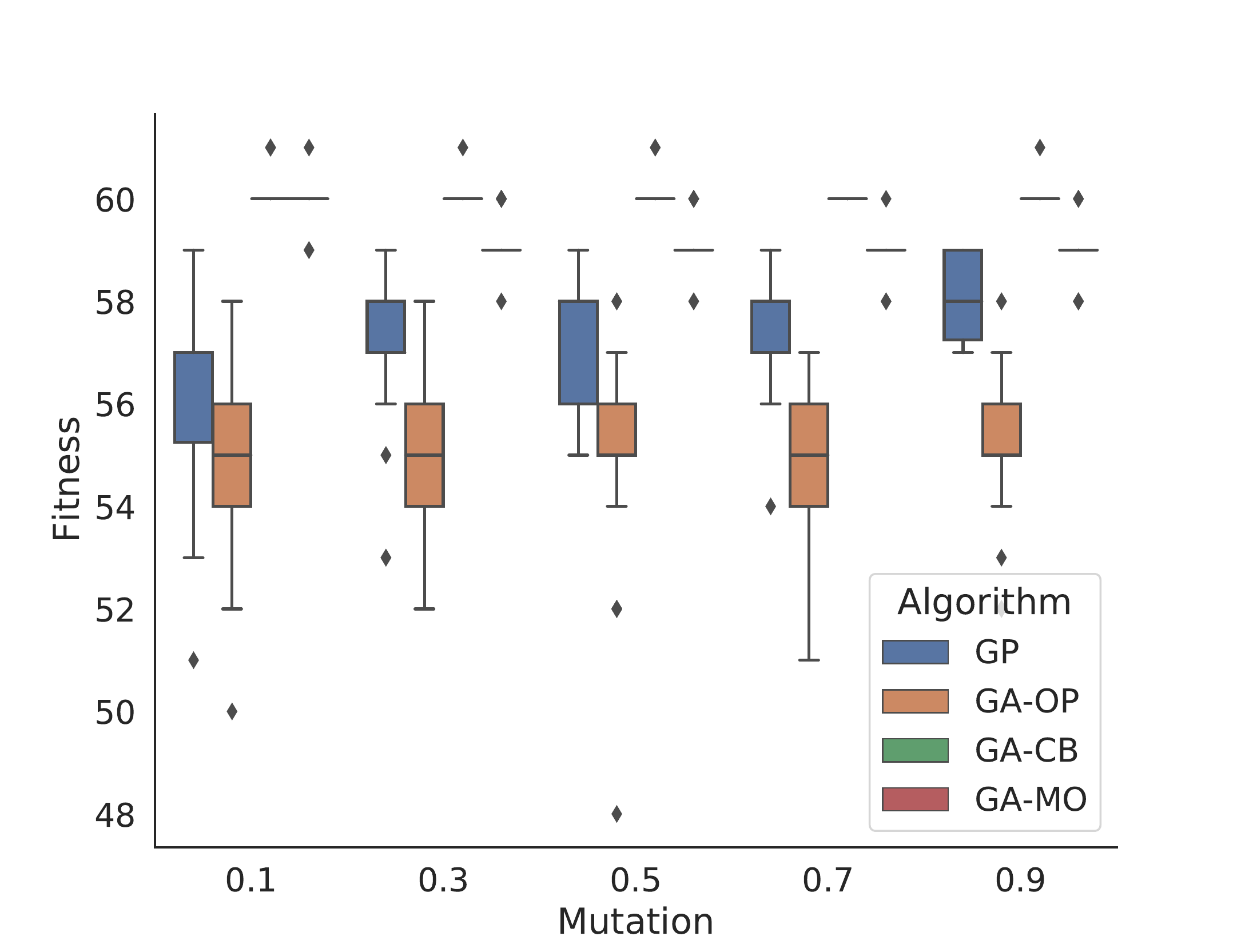}
\caption{Best fitness distributions of the four EAs for various mutation probabilities.}
\label{fig:tuning}
\end{figure}

Figure~\ref{fig:tuning} plots the distribution of the best fitness obtained by the four EA variants across all considered mutation rates.

The boxplots clearly show a stark difference in performances depending on the solutions encoding. In particular, the EAs based on the truth table representation generally behave worse than those exploiting the weightwise balanced encoding. Further, within those using the truth table encoding, GP generally has better a performance than GA-OP. On the other hand, there are no significant differences between the best fitness achieved by GA-CB and GA-MO, which indicates that the weightwise balanced representation plays a key role rather than the specific crossover operator employed. Finally, for each considered EA, there are no significant differences concerning the mutation rates, suggesting that the behavior of both GA and GP is robust with respect to this parameter. Thus, in our subsequent analysis, we selected the mutation rate yielding the highest number of occurrences of the maximum best fitness for each of the four considered EAs. This resulted in $p_{mut} = 0.1$ for GA-OP, GA-CB and GA-MO, and $p_{mut} = 0.9$ for GP.

\begin{figure}[t]
\centering
\includegraphics[width=\columnwidth]{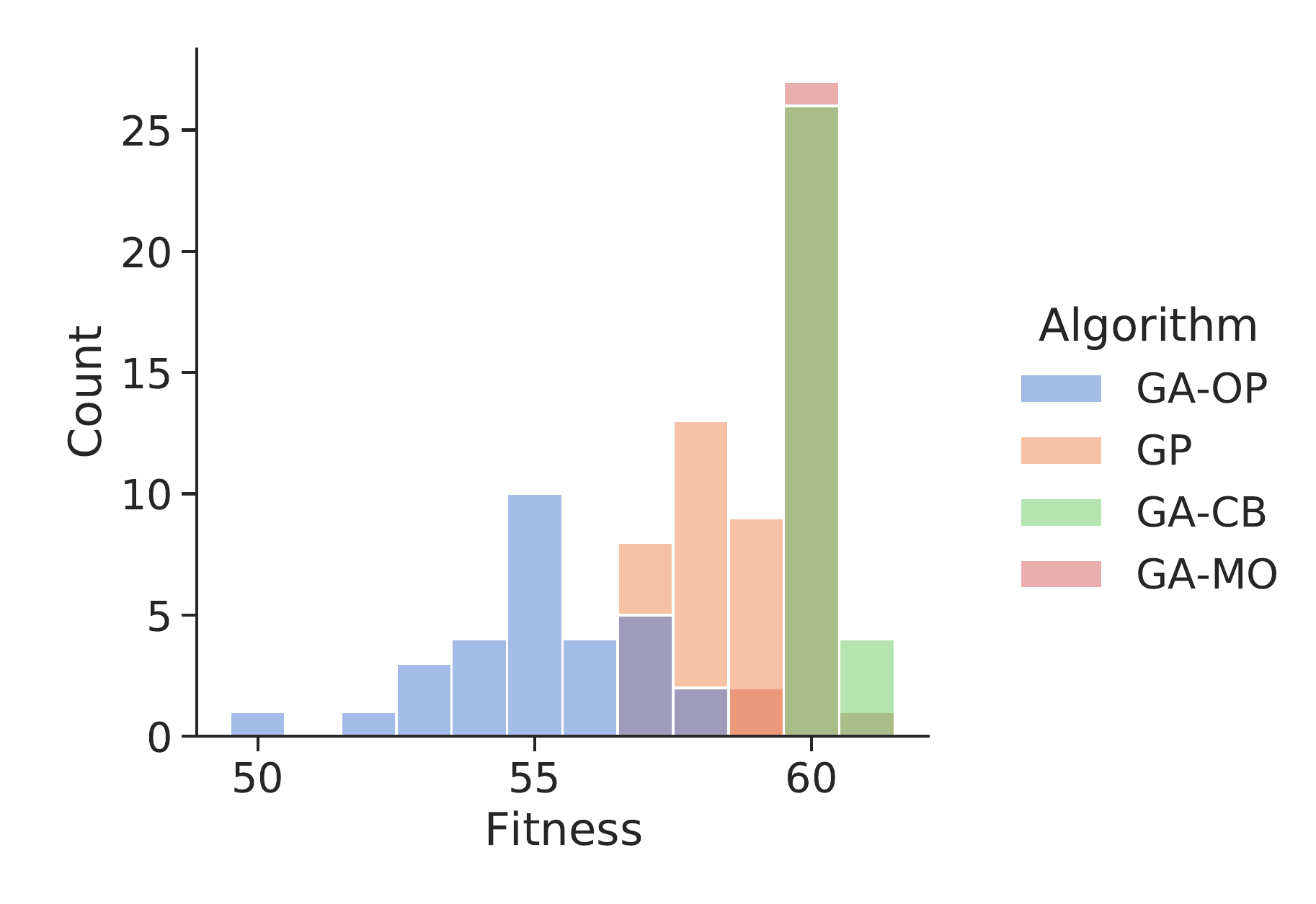}
\caption{Best fitness distributions for the four EA variants.}
\label{fig:barplot}
\end{figure}

Table~\ref{tab:results} summarizes the main statistical indicators for the best fitness obtained by the four algorithms according to the selected mutation rates.
\begin{table}[t]
    \centering
    \caption{Caption}
    \begin{tabular}{cccccc}
    \toprule
    Algorithm & Average & Std. Dev. & Median & Min & Max \\
    \midrule
    GA-OP & 55.07 & 1.80 & 55 & 50 & 58 \\
    GA-CB & 60.13 & 0.36 & 60 & 60 & 61 \\
    GA-MO & 59.97 & 0.32 & 60 & 59 & 61 \\
    GP    & 58.03 & 0.76 & 58 & 57 & 59 \\
    \bottomrule
    \end{tabular}
    \label{tab:results}
\end{table}
From the table, one can see better the difference between the EAs that respectively adopt the truth table the weightwise balanced representation. To show even more in detail such difference, in Figure~\ref{fig:barplot} we plot the distributions of the best fitness across all 30 experimental runs. Each bin reports the number of occurrences of the corresponding fitness value, with each EA represented by different colors. Combinations of the four key colors represent overlappings between distributions.
The performance gap becomes sharply evident by looking at these distributions. Indeed, one can see that GA-OP is the worst performing algorithm, having the widest dispersion of fitness values. Moreover, the most frequent best fitness for GA-OP is 55, which is the lower among all four considered EAs. GP scores a considerably better performance since its best fitness values range in the interval $[57,59]$, and the corresponding distribution has a lower standard deviation than GA-OP. Finally, the best performing variants are represented by the GA with the weightwise balanced representation, i.e., GA-CB and GA-MO, which both achieve the highest observed fitness values of $60$ and $61$. Moreover, the two EAs always converge to one of these two values, leading to the distributions with the lowest standard deviations. Although there are no significant differences between the two distributions, one can observe that the counter-based crossover is slightly better than the map-of-ones operator, since the GA evolved four solutions of fitness 61 with the former, as opposed to only one with the latter.

\section{Discussion}
\label{sec:discussion}

The most interesting remark arising from the results presented in the previous section is that \emph{GA was able to outperform GP} by using the weightwise balanced representation. This is somewhat surprising, as the empirical evidence gained so far in the relevant literature is that GP is usually better than GA when evolving the cryptographic properties of Boolean functions~\cite{picek16}. Previous authors linked this gap in performances to the underlying representation, with the GP trees likely having an advantage over the direct bitstring used by GA to encode the truth table of a Boolean function. As shown by Manzoni et al.~\cite{manzoni20}, the use of balanced operators such as counter-based and map-of-ones crossovers improves the GA performance over the classic one-point crossover. However, this improvement is still far from reaching the same GP results when evolving highly nonlinear balanced Boolean functions, especially for larger sizes.

On the opposite, the situation in the WPB functions problem addressed in this paper is reversed, with the weightwise balanced representation providing an advantage to GA, allowing it to score better fitness values than GP. We suspect that this is due to the highly constrained structure of the space of WPB functions. Indeed, one may argue that the improvement given by balanced crossover operators to GA when evolving globally balanced Boolean functions is no match for GP for a twofold reason. First, as we observed in Section~\ref{subsec:encoding}, the reduction in the size of the search space granted by the use of balanced operators is not really significant when compared to the space of all Boolean functions, with a difference of only an order of magnitude when $n=8$. Furthermore, one may also argue that minimizing the global unbalancedness of a generic Boolean function is a rather easy optimization objective for GP. Therefore, only a few fitness evaluations are needed before GP converges over a balanced solution. On the other hand, GP is dealing with several unbalancedness penalty factors for WPB functions, namely one for each Hamming weight $k$ between $1$ and $n-1$. This induces more constraints for the feasible solutions of the problem, and it might be the case that GP is wasting many fitness evaluations just to minimize the penalty factor, which leaves less room to optimize the sum of nonlinearities once a WPB function is reached.

\section{Conclusions and Future Work}
\label{sec:conclusions}
In this paper, we investigated the construction of weightwise perfectly balanced Boolean functions by means of GP and GA. Such functions recently became relevant in the design of stream ciphers based on the filter permutator paradigm, such as FLIP. Although the Boolean functions involved in those designs are defined over hundreds of input variables -- which prevents the use of any traditional metaheuristic to construct them --, the structure of the space of WPB functions is still largely unknown in general. This makes the use of EAs interesting to investigate the properties of WPB functions of small sizes. In particular, here, we considered functions of $n=8$ variables since it is the only problem instance where the restricted Walsh transforms can be computed using the naive method in a reasonable amount of time. We considered two different encodings for the candidate solutions, namely the classic truth table representation (largely used in other related works on EAs and Boolean functions) and the weightwise balanced representation. The latter stems from the observation that one can limit a GA to explore only among the space of WPB functions, which allows one to focus the optimization effort on maximizing the restricted nonlinearities. In this case, the GA can leverage on the use of balanced crossover operators~\cite{manzoni20} to preserve such encoding in the offspring solutions. In particular, these operators here need to be applied independently on each subset $E_{n,k}$ of inputs with Hamming weight $k$. We experimented with two fitness functions, one maximizing the sum of restricted nonlinearities and the other maximizing the minimum nonlinearity. When using the truth table representation, the fitness functions also optimized an unbalancedness penalty factor to converge on a WPB function.

Our results show that contrarily to the evidence gathered in related works on EA and Boolean functions, GA with balanced crossover operators achieve the best fitness over all experimental runs and outperforms by far GP with the truth table representation. We elaborated on this finding by observing that most of the previous work on the design of Boolean functions with EA focuses on the global balancedness property -- something which can be easily achieved by GP even with the classic tree-to-table representation. In this problem, the penalty factor minimized by GP instead requires a much more considerable optimization effort before obtaining a WPB function. On the other hand, a GA based on a weightwise balanced representation already starts from a population of WPB functions and can therefore concentrate only on maximizing the restricted nonlinearities.

There are several directions for future research on this optimization problem. A first idea would be to explore more in detail the potential of the weightwise balanced representation in designing WPB functions with high nonlinearity profiles. For example, it could be interesting to apply partially balanced crossover operators such as the ``tip the balance'' strategy proposed in~\cite{manzoni20a}. Additionally, we believe it would be interesting to explore whether it is possible to evolve secondary constructions of WPB Boolean functions for future work. While EAs can evolve such Boolean function, we see a problem with scalability due to the computation cost of calculating the Walsh-Hadamard spectrum with a naive approach. Thus, having constructions that generalize to any input size does seem the best option. 

\bibliographystyle{abbrv}
\bibliography{bibliography}

\end{document}